%% file: main.tex
\documentclass[letterpaper]{article} 
\usepackage{aaai2027}
\nocopyright
\usepackage[hyphens]{url}  
\usepackage{graphicx} 
\usepackage{natbib}  
\usepackage{caption} 
\usepackage{booktabs}
\usepackage{colortbl}
\usepackage{amsmath}
\usepackage{amssymb}
\usepackage{etoolbox}
\usepackage{algorithm}
\usepackage{algorithmic}
\usepackage{array}
\makeatletter
\patchcmd{\@maketitle}
  {Corresponding author\ifaaai@corrmulti{s}\fi.}
  {Corresponding authors.}
  {}{\PackageWarning{main}{Could not patch corresponding-author footnote}}
\makeatother

\definecolor{tablegray}{gray}{0.94}

\newcommand{\pdfanchorlabel}[1]{\label{#1}}
\newcommand{\Figref}[1]{Figure~\ref{#1}}
\newcommand{\Tabref}[1]{Table~\ref{#1}}

\title{Multi-View Unified Camera Fields: Geometry-Shaped Action-Facing Representations for RGB-Only Multi-Camera VLA Policies}
\author{
Jiarui Yang\textsuperscript{\rm 1}, Yehao Lu\textsuperscript{\rm 2}, Yuning Su\textsuperscript{\rm 3}, Yufeng Xie\textsuperscript{\rm 4}, Yu Zhong\textsuperscript{\rm 4}, Haiyu Lan\textsuperscript{\rm 4},\\
Tianjing Hao\textsuperscript{\rm 5}, Kaixiang Lu\textsuperscript{\rm 4}, Peiwen Lin\textsuperscript{\rm 4}, Chuang Wang\textsuperscript{\rm 4}, Enyu Li\textsuperscript{\rm 4}\corresponding, Junwei Liang\textsuperscript{\rm 1}\corresponding
}
\affiliations{
\textsuperscript{\rm 1}The Hong Kong University of Science and Technology (Guangzhou),
\textsuperscript{\rm 2}Zhejiang University,\\
\textsuperscript{\rm 3}Simon Fraser University,
\textsuperscript{\rm 4}Agibot,
\textsuperscript{\rm 5}Xi'an Jiaotong University
}

\begin{document}

\maketitle

\begin{abstract}
Vision-Language-Action (VLA) models have shown strong generalization in robotic manipulation, yet complex contact-rich tasks often benefit from multi-camera observations that jointly capture the end effector, objects, and targets under occlusion. Existing multi-camera VLAs usually concatenate view tokens, leaving action representations weak in metric depth and inconsistent across cameras. We introduce \emph{Multi-View Unified Camera Fields} (MVUCF), a training-only framework that forms a shared action-facing latent field across views. A coordinate-query depth objective makes metric depth recoverable, while a preprocessing-aware correspondence objective aligns tokens observing the same physical point from different cameras. Both directly shape the hidden states consumed by the action module. After geometry injection, depth, camera calibration, and auxiliary heads are removed, so deployment uses the original RGB-only graph with no extra inference FLOPs. Held-out probes confirm stronger depth recovery and cross-view matching. Under matched GR00T-N1.6 settings, MVUCF reaches 98.9\% on LIBERO, improves LIBERO-Plus by 22.4 points, and raises success by 23.3 points across six RoboTwin tasks spanning three action families: touch, move-and-place, and contact interaction. Real-world humanoid experiments further provide evidence of its practical effectiveness under RGB-only deployment.
\end{abstract}

\begin{figure*}[t]
\centering
\includegraphics[width=\textwidth]{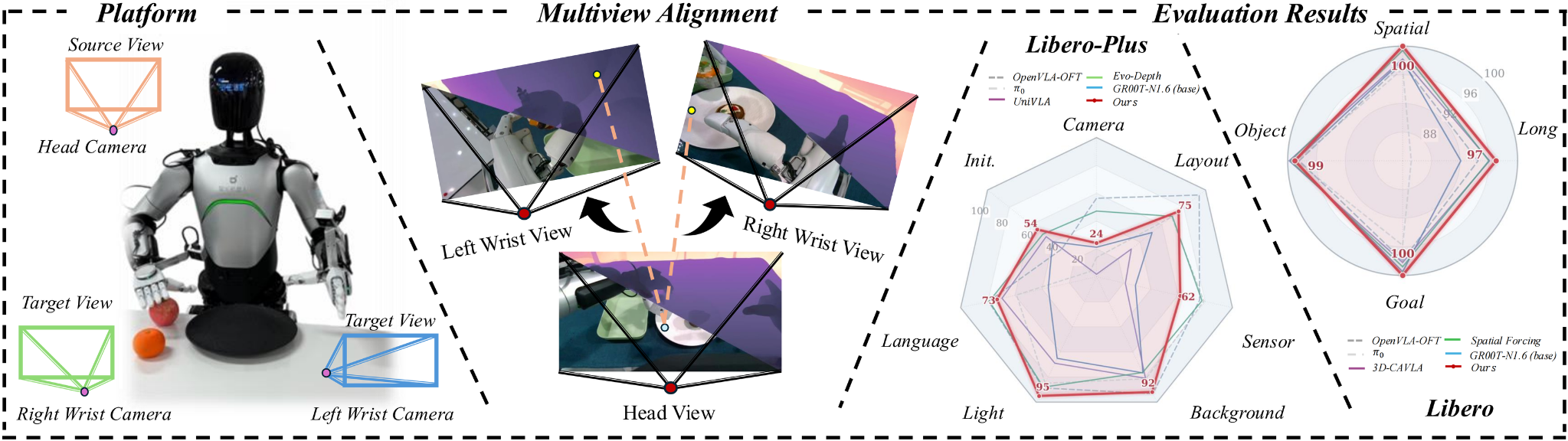}
\caption{Overview of \emph{Multi-View Unified Camera Fields}. Calibrated head-and-wrist views provide complementary observations; training-time depth and reprojection supervision reshape action-facing tokens, while deployment remains RGB-only with the native inference graph.}
\pdfanchorlabel{fig:teaser}
\end{figure*}

\section{Introduction}

Vision-language-action (VLA) models combine pretrained vision-language backbones with learned action generators and have demonstrated strong generalization across diverse robotic manipulation tasks~\cite{brohan2023rt2,kim2024openvla,octo2024,black2025pi0,nvidia2025gr00t}. Contact-rich manipulation, however, places stricter demands on visual perception. A single camera may fail to preserve fine-grained details of the end effector while simultaneously capturing the object, target, and their spatial relations, especially under occlusion. Multi-camera observations are therefore commonly introduced to provide complementary views of the workspace. Most existing multi-camera policies simply concatenate per-view visual representations and rely on sparse action supervision to infer how those views are physically related~\cite{zhao2023act,octo2024,nvidia2025gr00tn16}. Despite its simplicity, this strategy has two important limitations. First, because many VLA backbones are initialized from vision-language models pretrained primarily on 2D semantic data, their hidden states do not reliably recover the metric depth and 3D spatial structure contained in multi-camera observations~\cite{li2025spatialforcing}. Second, concatenation alone does not explicitly encourage features corresponding to the same physical point to remain consistent across cameras; more broadly, visual foundation models have been shown to represent the same surface inconsistently across views~\cite{elbanani2024probing}. \Figref{fig:main_diagnostic} provides direct evidence of both limitations: native backbone states exhibit weak metric-depth recovery and nearly random cross-view retrieval, even when observations from multiple cameras are available.

\begin{figure*}[t!]
\centering
\includegraphics[width=0.96\textwidth]{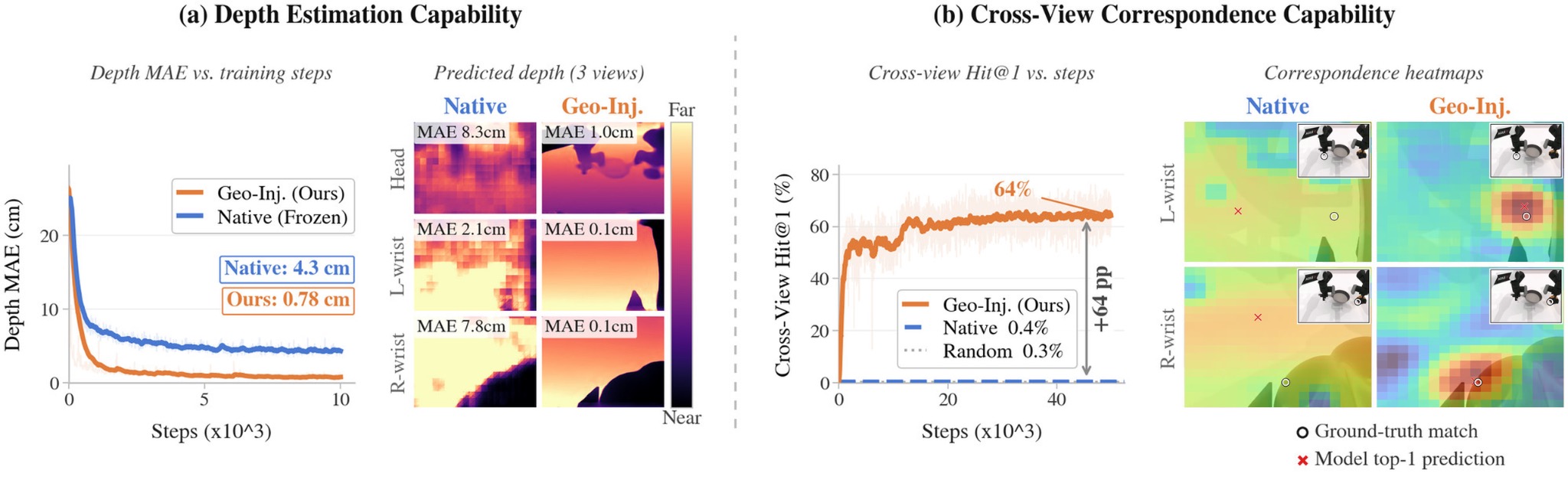}
\caption{Geometry-injection diagnostics. (a) With the same depth-head probe, metric depth is difficult to extract from the native backbone (4.3\,cm MAE) but becomes readily recoverable after geometry injection (0.78\,cm); the panels show per-view MAE for one frame. (b) Native tokens yield near-random cross-view retrieval (0.4\% Hit@1 versus 0.3\% random), indicating weak same-world-point consistency, whereas geometry injection reaches 64\% and localizes the projected target. Together, metric depth and cross-view token consistency---two prerequisites for constructing a multi-camera field---become accessible in the backbone.}
\pdfanchorlabel{fig:main_diagnostic}
\end{figure*}

Prior work has addressed these limitations largely in isolation. Geometry-aware policies either introduce explicit 3D cues or transfer geometric priors from models such as DUSt3R and VGGT into VLA representations~\cite{shridhar2022peract,wang2024dust3r,wang2025vggt,qu2025spatialvla,li2025spatialforcing,peng2026g3vla}. Meanwhile, Selfi improves cross-view consistency through reprojection-based feature alignment~\cite{deng2025selfi}. However, these methods mainly address separate prerequisites for constructing a unified multi-camera field: consistent representations across views and metric 3D grounding. Although these approaches make substantial contributions, their backbone-shaping mechanisms and representation targets differ substantially, preventing a straightforward modular combination. A more unified framework is therefore needed to simultaneously recover cross-view consistency and 3D structure in backbone representations, enabling VLA policies to construct an action-facing multi-view camera field.

We introduce \emph{Multi-View Unified Camera Fields} (MVUCF), a training-only framework that jointly injects these two properties into the token lattice consumed by the action module. During geometry injection, a coordinate-query depth objective makes metric surface distance recoverable at continuous grid locations. In parallel, a preprocessing-aware correspondence objective reprojects calibrated points between raw camera views, maps them through the exact backbone preprocessing transformation, and aligns tokens observing the same physical point. The two objectives provide complementary supervision: depth establishes within-view metric location, whereas correspondence establishes cross-view physical-point identity. Both objectives directly update the upper backbone hidden states that later feed the action module, forming a shared action-facing latent field across cameras rather than two disconnected auxiliary representations. After geometry injection, the auxiliary heads, depth observations, and camera calibration are removed. The geometry-shaped backbone is then frozen, and the native action module is trained on the resulting representations. Deployment consequently retains the original RGB-only inference graph without additional sensors, modules, or inference FLOPs.

In summary, our work makes the following key contributions to geometry-aware multi-camera VLA learning:

\noindent (i) We present a token-level diagnosis showing that native multi-camera VLA representations can remain weak in metric-depth recovery and cross-view correspondence, and that deep external geometry features need not preserve the exact token identity required for action grounding.

\noindent (ii) We introduce MVUCF, a unified training-only framework that jointly injects continuous metric-depth recoverability and same-world-point consistency into the action-facing VLA grid via a coordinate-query depth objective and a preprocessing-aware raw-camera-to-token correspondence objective.

\noindent (iii) We provide controlled evidence from held-out representation probes, matched-policy comparisons across LIBERO, LIBERO-Plus, and six RoboTwin tasks spanning three action families, component ablations, and real-world humanoid experiments, demonstrating that the jointly shaped representation improves multi-camera manipulation without requiring inference-time geometry.

\begin{figure*}[t!]
\centering
\includegraphics[width=\textwidth]{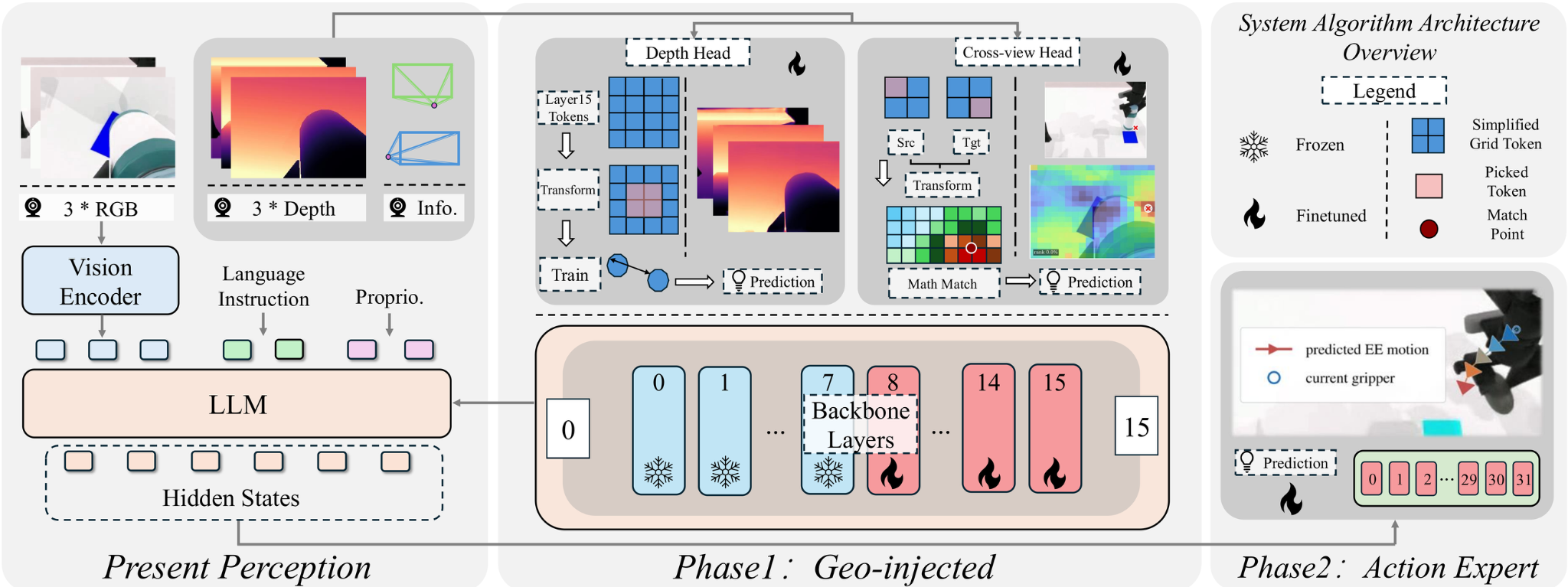}
\caption{System architecture of the proposed Multi-View Unified Camera Fields framework. During geometry injection, the backbone receives multi-camera RGB observations and layers 8--15 are updated by coordinate-based depth and cross-view matching objectives read from the layer-15 hidden grid. After injection, both heads are removed and the action head is trained on the geometry-injected hidden states, preserving the original inference structure.}
\pdfanchorlabel{fig:architecture}
\end{figure*}

\section{Related Work}

\noindent\textbf{VLA policies and spatial grounding.} RT-2, OpenVLA, Octo, $\pi_0$, and GR00T demonstrate broad vision-language-conditioned control~\cite{brohan2023rt2,kim2024openvla,octo2024,black2025pi0,nvidia2025gr00t}. Their semantic competence, however, does not guarantee metric depth or agreement between wrist and global views. Since broad vision-language backbone adaptation can alter pretrained semantic priors~\cite{hancock2025actions}, we localize geometry shaping to action-facing upper blocks while freezing the visual encoder and lower blocks.

\noindent\textbf{3D information injection.} Existing methods incorporate geometry into robotic policies either explicitly or implicitly. Explicit approaches expose the policy to RGB-D observations or calibrated 3D encodings: PerAct voxelizes RGB-D inputs, while camera-aware VLAs introduce 3D positions, viewing rays, or projective cues~\cite{shridhar2022peract,qu2025spatialvla,peng2026g3vla}. Such designs may require depth or calibration at inference and modify the original visual interface. In contrast, Spatial Forcing preserves RGB-only deployment by aligning VLA states with pretrained geometry features~\cite{li2025spatialforcing}. Such feature matching transfers a geometric prior but directly enforces neither metric-depth recoverability nor same-point correspondence on the action-facing grid. VGGT also does not guarantee cross-view token consistency: \Figref{fig:vggt_diagnostic} shows same-point discriminability degrading in deeper layers. We therefore supervise both properties directly on action-facing VLA states. We regard publicly available 2026 arXiv manuscripts as concurrent work.

\noindent\textbf{Multi-view fusion.} El Banani et al. show that visual foundation models struggle to maintain consistent surface representations across views, especially under large viewpoint changes~\cite{elbanani2024probing}. Selfi addresses this issue in 3D reconstruction by using predicted depth and camera parameters to train a reprojection-aligned feature adapter~\cite{deng2025selfi}. Nevertheless, the former diagnoses multi-view inconsistency without proposing a fusion mechanism, while the latter is designed for reconstruction, novel-view synthesis, and pose estimation rather than action-facing VLA representations. Neither directly establishes a control-oriented representation that jointly preserves same-point consistency and metric geometry.

\section{Method}

\subsection{Overview}

As illustrated in \Figref{fig:architecture}, we build on GR00T-N1.6~\cite{nvidia2025gr00t,nvidia2025gr00tn16}. Given synchronized RGB observations $\mathcal{O}_t=\{\mathbf{o}_t^i\}_{i=1}^{V}$ and a language instruction $l$, the policy predicts
\begin{equation}
    \mathbf{a}_t \sim \pi_\theta(\cdot\mid \mathcal{O}_{\leq t}, l).
\end{equation}

Before action training, we first inject geometry into the backbone using the official demonstrations together with synchronized depth and camera calibration. Depth supervision makes within-view metric structure recoverable, while cross-view supervision aligns tokens that observe the same physical point from different cameras. Because both objectives operate on the same action-facing feature grid later consumed by the policy, geometric information is incorporated directly into the backbone representation rather than stored in a separate auxiliary branch.

To construct accurate supervision, projected target locations are kept continuous and mapped through the complete image-preprocessing pipeline before being associated with backbone tokens. This avoids the coordinate mismatch that would arise from directly scaling raw-camera pixels to feature-grid indices. After geometry injection, the depth and cross-view heads are removed, and the geometry-shaped backbone is frozen. The native action module is then trained using RGB observations only. Consequently, both action training and deployment retain the original RGB-only inference graph without requiring depth, camera calibration, or auxiliary geometry modules.

\subsection{Cross-View Geometry and Target Construction}
\label{sec:target_construction}

Labels are constructed in calibrated raw-camera coordinates before being mapped to the backbone grid. For source pixel $\mathbf u_i^{\mathrm{raw}}$ with depth $z_i^\ast$, intrinsics $\mathbf K_i$, and camera-to-world pose $\mathbf T_{i\to w}$, we back-project the pixel into 3D and reproject it into target view $j$:
\begin{equation}
\label{eq:raw_reprojection}
\begin{split}
\mathbf P_w &= \mathbf T_{i\to w}\,\pi^{-1}(\mathbf u_i^{\mathrm{raw}},z_i^\ast,\mathbf K_i),\\
\mathbf u_j^{\mathrm{raw}} &= \pi(\mathbf T_{w\to j}\mathbf P_w,\mathbf K_j).
\end{split}
\end{equation}
As illustrated in \Figref{fig:method_crossview_logic}, each reprojected target must lie within the image bounds, have a valid target-frame depth in $[0.05,5.0]$ m, and satisfy the z-buffer consistency condition $|Z_j-z_j^\ast(\mathbf u_j^{\mathrm{raw}})|<0.10$ m. These checks remove invalid, occluded, or depth-inconsistent correspondences before the accepted pixel is mapped to the positive token center $\mathbf q_j^\ast=T_{\mathrm{img}\to\mathrm{grid}}(\mathbf u_j^{\mathrm{raw}})$ and its Gaussian target. Here, $T_{\mathrm{img}\to\mathrm{grid}}$ composes GR00T's letterboxing, resizing, cropping, smart-resize, patchification, and pixel-unshuffle operations. This project-before-preprocess order avoids the label error caused by naively scaling raw pixels to token coordinates. Camera calibration is used only to construct training targets and is absent during action training and deployment.

\begin{figure}[t]
\centering
\includegraphics[width=0.98\columnwidth]{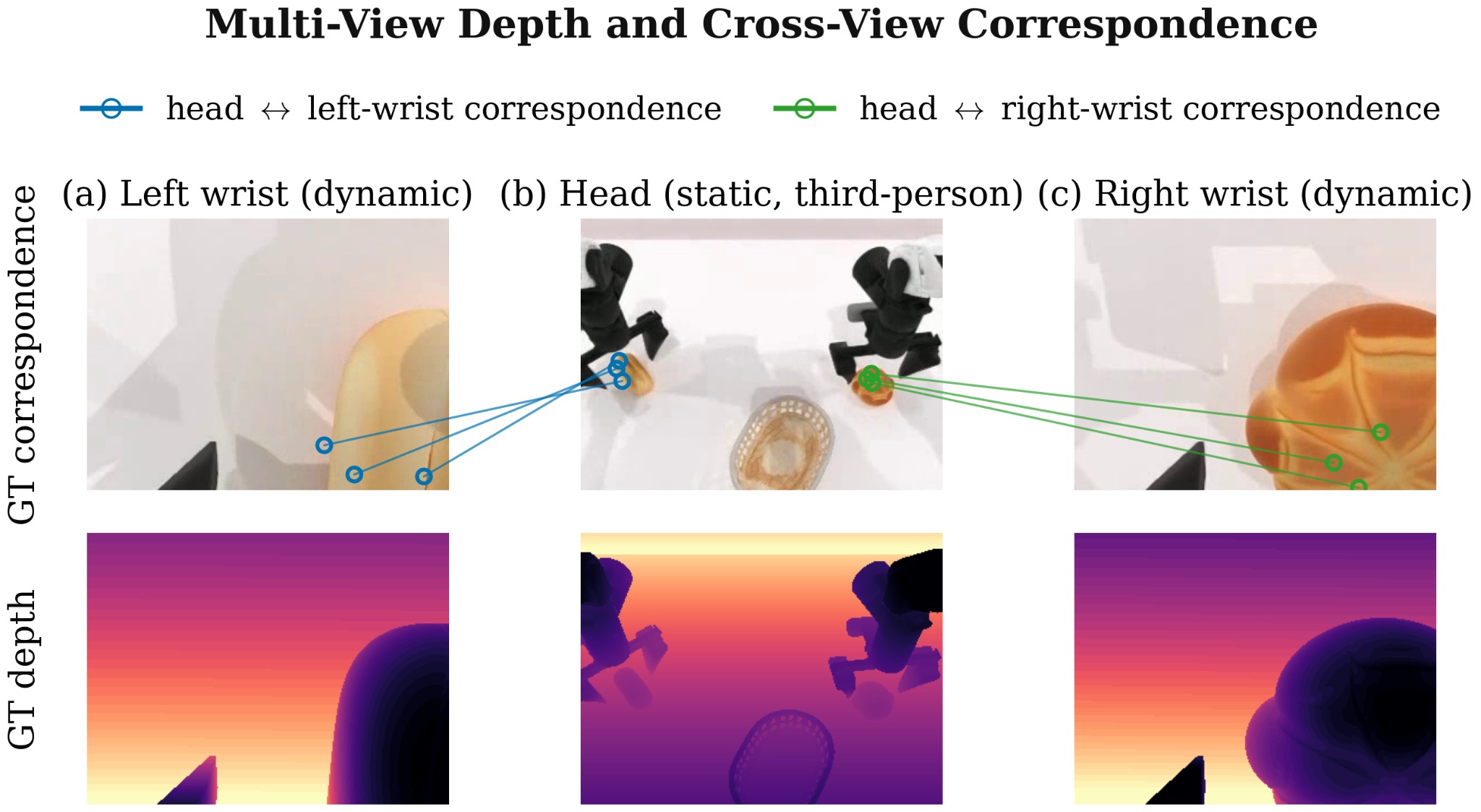}
\caption{Cross-view target construction. Depth lifts a source pixel to 3D, calibrated poses reproject the same point, and the exact backbone preprocessing maps the accepted raw pixel to its positive token center.}
\pdfanchorlabel{fig:method_crossview_logic}
\end{figure}

\subsection{Coordinate-Query Depth Head}

For each view $i$, the depth head predicts metric depth at a continuous grid coordinate $\mathbf q=(x,y)$:
\begin{equation}
    \hat{z}_i(\mathbf{q}) = g_\phi(\mathbf{F}_t^i, \mathbf{q}).
\end{equation}
Each view supplies 3072 valid queries, with half sampled uniformly and half concentrated on nearby surfaces and depth discontinuities. A grid-preserving trunk (LN$_{2048}$--Linear$_{1024}$--GELU--Conv$_{3\times3,1024}$--GELU) is applied before bilinear feature lookup. The depth head concatenates the sampled 1024-dimensional feature, local $x/y$ differences, and the sub-cell phase $\boldsymbol\varphi(\mathbf q)=\mathbf q-\lfloor\mathbf q\rfloor$, forming a 3074-dimensional query representation without an absolute-position input. An MLP$_{3074\to1024\to1024\to1}$ with Softplus predicts metric depth, while a parallel MLP$_{3074\to1024\to1}$ predicts log variance for uncertainty-aware supervision. The depth target is obtained through nearest-neighbor lookup $D_i[\operatorname{round}(T_{\mathrm{grid}\to\mathrm{raw}}(\mathbf q))]$, with invalid target locations masked.

The depth objective is
\begin{equation}
\begin{aligned}
\mathcal{L}_{\mathrm{depth}}={}&0.5\mathcal{L}_{\mathrm{silog}}+1.25\mathcal{L}_{\mathrm{inv}}+1.0\mathcal{L}_{\mathrm{grad}}\\
&+0.1\mathcal{L}_{\mathrm{seam}}+0.05\mathcal{L}_{\mathrm{unc}}.
\end{aligned}
\end{equation}
Inverse-depth supervision emphasizes the nearby manipulation workspace, while neighboring $+x$ and $+y$ queries preserve local surface variation. All objective weights and query-sampling strategies are fixed across datasets; Appendix~\ref{app:metric_grounding} provides the complete definitions of each loss term and the sampling procedure.

\subsection{Cross-View Correspondence Head}
\label{sec:cv_matching}

A shared LN$_{2048}$--Linear$_{2048\to128}$ projector followed by $L_2$ normalization produces source and target embeddings $\mathbf e_m^s$ and $\mathbf e_n^t$. Their global matching logits are
\begin{equation}
\label{eq:cv_logits}
\ell_{mn}=10(\mathbf e_m^s)^\top\mathbf e_n^t.
\end{equation}
For target-token center $\mathbf c_n$, the soft correspondence target is $y_{mn}\propto\exp(-\|\mathbf c_n-\mathbf q_j^\ast\|_2^2/(2\sigma^2))$ with $\sigma=0.75$. This distribution supervises a spatial neighborhood around the projected point rather than assigning all probability to a single discretized grid cell. Hard-negative InfoNCE draws spatial-ring, in-batch, and alternate-view negatives, while a margin loss separates the projected match from the strongest non-positive candidate. A $3\times3$ local branch is trained to resolve sub-token neighborhoods, with its logit scale annealed during training.

The final correspondence objective is
\begin{equation}
\mathcal L_{\mathrm{cv}}=\mathcal L_{\mathrm{softCE}}+0.1\mathcal L_{\mathrm{hnce}}+0.1\mathcal L_{\mathrm{margin}}.
\end{equation}
With margin $\gamma=0.2$, the outer weight is annealed during training, emphasizing same-point alignment early and relaxing the auxiliary constraint before action learning. Because correspondence targets and matching logits share the same post-preprocessing layer-15 lattice, the head directly shapes the backbone state later consumed by the action module. Spatially softened positives absorb sub-token projection error, while spatial-ring negatives prevent nearby but incorrect patches from collapsing into equivalent matches. The resulting objective preserves continuity across neighboring grid cells without weakening discrimination among nearby candidate locations.

\section{Experiments}

\subsection{Evaluation protocol}
We organize evaluation as one evidence chain: latent-property diagnostics; RGB-only control on LIBERO~\cite{liu2023libero}, LIBERO-Plus~\cite{fei2025liberoplus}, and six-task RoboTwin~\cite{mu2025robotwin}; controlled ablation; and a physical humanoid pilot. Base and Ours share action data, head, schedule, checkpoint rule, and inference graph; geometry injection is the only controlled difference. Published systems provide context, while attribution rests on the matched GR00T pair. LIBERO SDs are computed across training seeds; other intervals quantify fixed-checkpoint rollout uncertainty. Full protocols are provided in Appendix~\ref{app:evaluation}. We report SR, depth MAE, Hit@1, and retrieval rank. For true-match similarities $A$ and random-pair similarities $B$, Cohen's effect size is computed as
\begin{equation}
\label{eq:cohen_d}
d(A,B)=\frac{\mu_A-\mu_B}{\sqrt{(\sigma_A^2+\sigma_B^2)/2}}.
\end{equation}

\paragraph{Training and deployment cost.}
All geometry-injection and action-training runs use a global batch size of 128. Geometry injection runs for 50k steps and takes approximately 18 wall-clock hours. Action schedules are benchmark-specific: 60k steps for LIBERO and 180k joint steps for the six RoboTwin tasks (30k task-equivalent steps each). We use pre-specified terminal checkpoints without downstream-success selection. Geometry-specific inputs and auxiliary heads are removed before action training, so this offline stage leaves the native RGB-only inference graph and runtime unchanged.

\subsection{Representation diagnostics}

\begin{figure}[t]
\centering
\includegraphics[width=0.94\columnwidth]{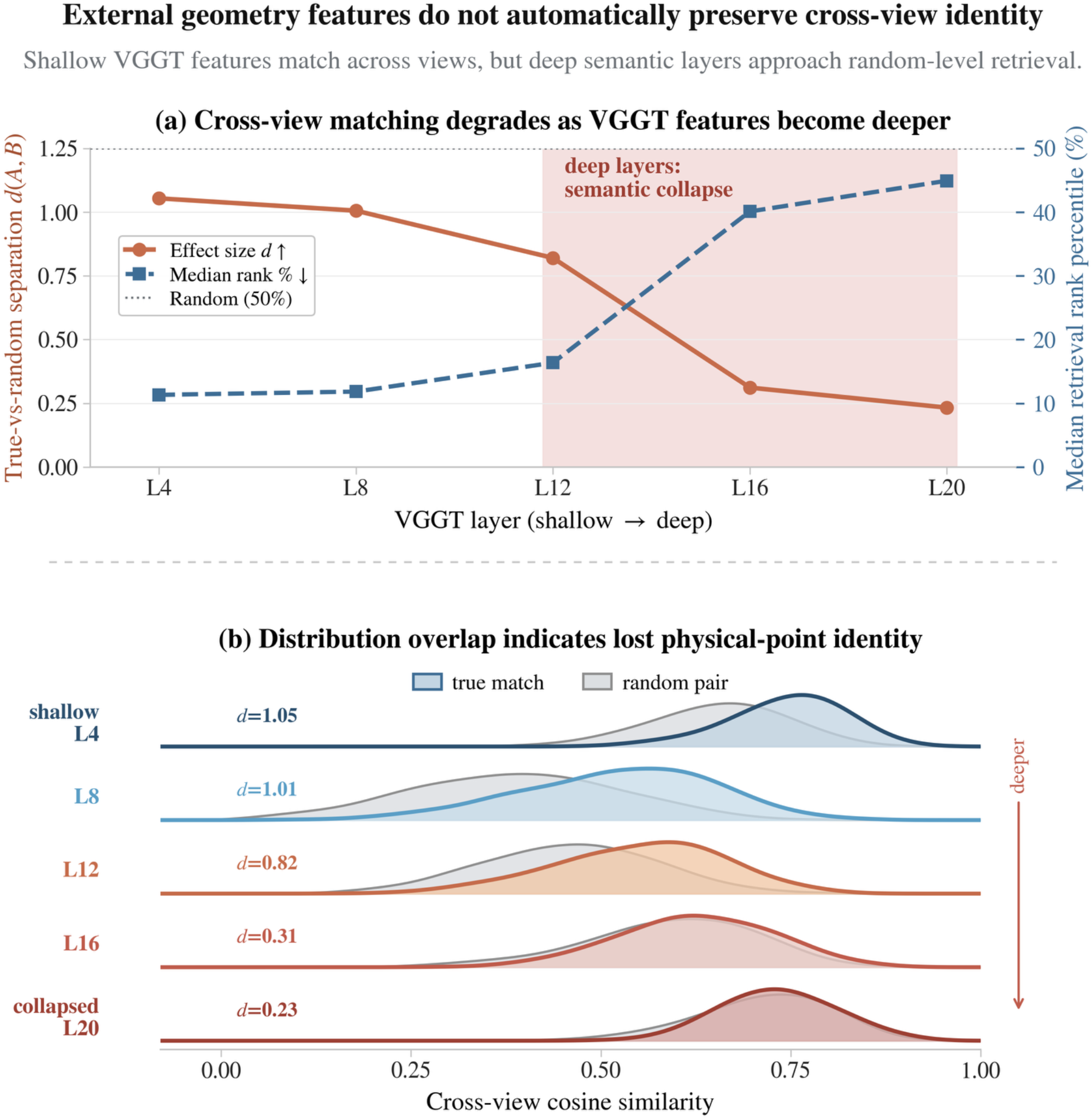}
\caption{Layer-wise VGGT diagnostic. Shallow features preserve cross-view token identity, whereas deeper features approach random retrieval after global attention mixing. Larger $d(A,B)$ and lower median rank percentile are better.}
\pdfanchorlabel{fig:vggt_diagnostic}
\end{figure}

\paragraph{Is a deep external geometry feature sufficient?}
\Figref{fig:vggt_diagnostic} evaluates whether a single frozen VGGT layer would provide an adequate distillation target. Shallow VGGT features separate true matches, whereas deeper features lose token-level discriminability after global attention mixing. Scene-level 3D abstraction therefore need not preserve the exact patch identity required for contact localization. This result motivates explicit metric and correspondence supervision on the action-facing VLA grid rather than treating one deep external layer as a unified camera field.

\paragraph{Does geometry injection reshape the VLA states?}
\Figref{fig:geo_results} evaluates the geometry-shaped representations on held-out data. After geometry injection, true correspondences separate clearly from random pairs, similarity concentrates around the projected target, and metric depth becomes substantially more recoverable. The depth and retrieval probes measure complementary properties---within-view metric structure and same-point identity across cameras---and improve consistently across effect size, retrieval rank, depth MAE, and the fraction of predictions within 2\,cm. Their agreement therefore demonstrates that geometry injection produces the intended representation change rather than merely reducing a training objective.

\begin{figure*}[t!]
\centering
\includegraphics[width=0.92\textwidth]{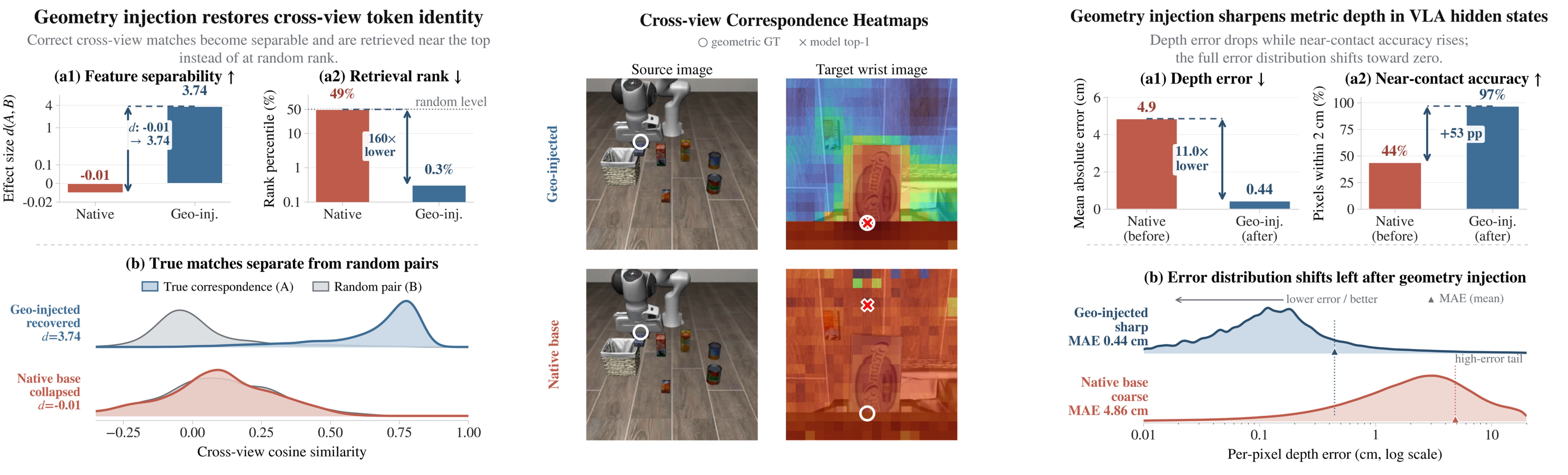}
\caption{Full-evaluation diagnostics. Unlike the single-frame per-view visualization in \Figref{fig:main_diagnostic}, the results here report full-evaluation mean depth MAE. Geometry injection separates true cross-view matches, localizes projected correspondences, reduces mean depth MAE from 4.9 to 0.44\,cm, and raises predictions within 2\,cm from 44\% to 97\%.}
\pdfanchorlabel{fig:geo_results}
\end{figure*}

\paragraph{Does the change reach the action pathway?}
At matched RoboTwin rollout steps, \Figref{fig:attn_analysis} shows that geometry injection concentrates action-to-image attention more tightly around the manipulated object and target. Because the panels are normalized independently, the comparison concerns spatial localization rather than absolute attention magnitude. Together with the held-out representation probes and task-level improvements, this localized pattern provides complementary evidence that the geometry-shaped representation propagates into the action pathway.

\begin{figure}[t]
\centering
\includegraphics[width=0.96\columnwidth]{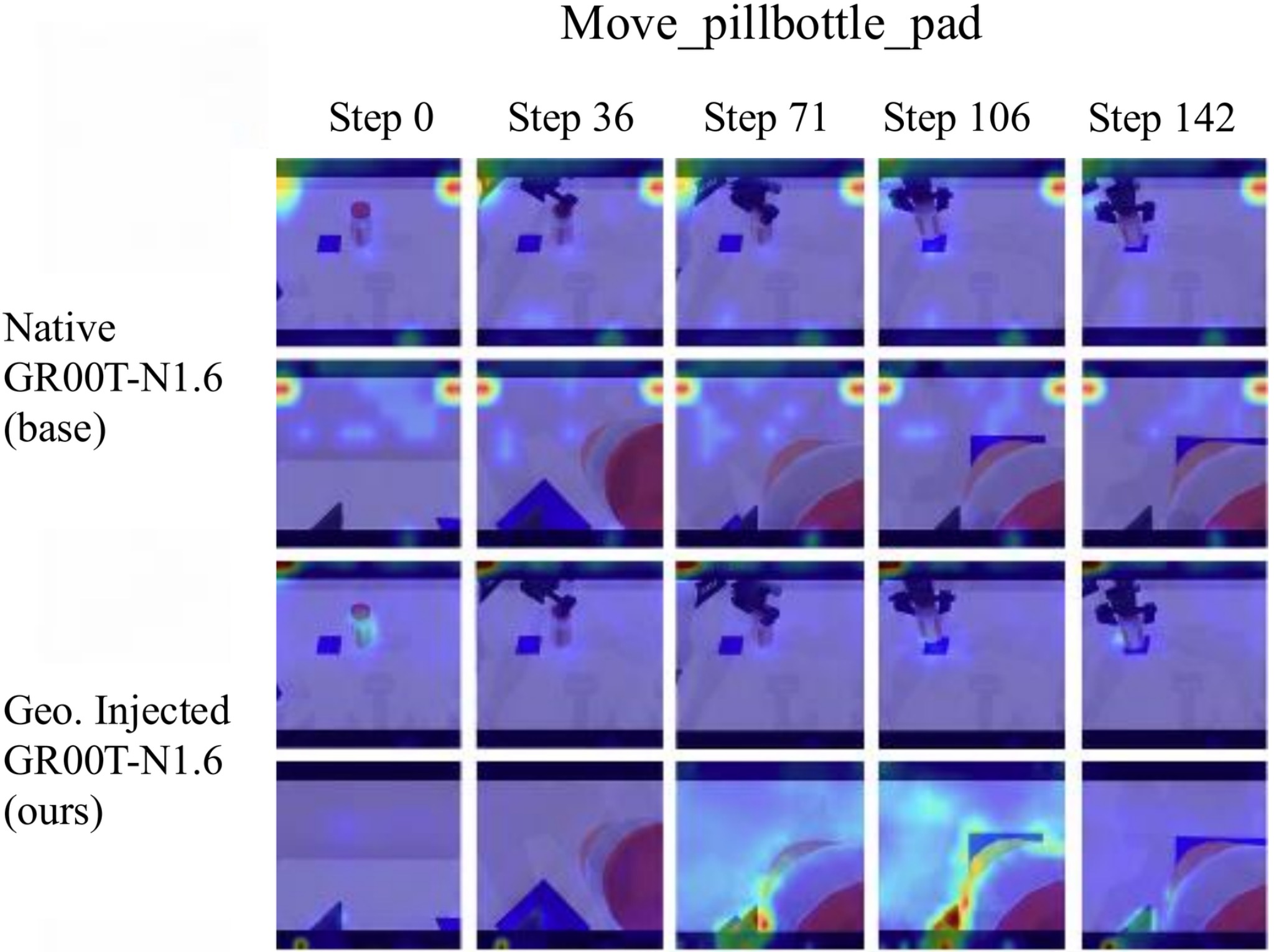}
\caption{Action-to-image cross-attention for \texttt{move\_pillbottle\_pad} at matched rollout steps. Independent normalization highlights the spatial support within each panel; the comparison concerns localization rather than absolute attention magnitude. Attention extraction details are provided in Appendix~\ref{app:diagnostics}.}
\pdfanchorlabel{fig:attn_analysis}
\end{figure}

\subsection{Simulation evaluation}

\paragraph{1) Standard LIBERO.}
\begin{table*}[t!]
\centering
{\small
\setlength{\tabcolsep}{4.2pt}
\renewcommand{\arraystretch}{1.02}
\begin{tabular}{lccccc}
\toprule
Method & Spatial SR (\%) & Object SR (\%) & Goal SR (\%) & Long SR (\%) & Average SR (\%) \\
\midrule
\rowcolor{tablegray}
\multicolumn{6}{c}{\textbf{2D VLA}} \\
OpenVLA~\cite{kim2024openvla} & 84.7 & 88.4 & 79.2 & 53.7 & 76.5 \\
Octo~\cite{octo2024} & 78.9 & 85.7 & 84.6 & 51.1 & 75.1 \\
$\pi_0$~\cite{black2025pi0} & 96.8 & 98.8 & 95.8 & 85.2 & 94.2 \\
Diffusion Policy~\cite{chi2023diffusion} & 78.3 & 92.5 & 68.3 & 50.5 & 72.4 \\
TraceVLA~\cite{zheng2025tracevla} & 84.6 & 85.2 & 75.1 & 54.1 & 74.8 \\
Dita~\cite{hou2025dita} & 84.2 & 96.3 & 85.4 & 63.8 & 82.4 \\
CoT-VLA~\cite{zhao2025cotvla} & 87.5 & 91.6 & 87.6 & 69.0 & 83.9 \\
$\pi_0$-FAST~\cite{pertsch2025fast} & 96.4 & 96.8 & 88.6 & 60.2 & 85.5 \\
UniVLA~\cite{bu2025latentactions} & 96.5 & 96.8 & 95.6 & 92.0 & 95.2 \\
OpenVLA-OFT~\cite{kim2025oft} & 97.6 & 98.4 & 97.9 & 94.5 & 97.1 \\
\midrule
\rowcolor{tablegray}
\multicolumn{6}{c}{\textbf{Implicit 3D VLA}} \\
Spatial Forcing~\cite{li2025spatialforcing} & \underline{99.4} & \underline{99.6} & \underline{98.8} & 96.0 & \underline{98.5} \\
\midrule
\rowcolor{tablegray}
\multicolumn{6}{c}{\textbf{Explicit 3D VLA}} \\
SpatialVLA~\cite{qu2025spatialvla} & 88.2 & 89.9 & 78.6 & 55.5 & 78.1 \\
GeoVLA~\cite{sun2025geovla} & 98.4 & 99.0 & 96.6 & \underline{96.6} & 97.7 \\
3D-CAVLA~\cite{bhat2025cavla} & 98.2 & \textbf{99.8} & 98.2 & 96.1 & 98.1 \\
\midrule
\rowcolor{tablegray}
\multicolumn{6}{c}{\textbf{GR00T-based comparison}} \\
GR00T-N1.6 (Base)~\cite{nvidia2025gr00tn16} & 99.3$\pm$1.2 & 99.2$\pm$0.8 & 98.4$\pm$1.2 & 92.9$\pm$1.0 & 97.4$\pm$0.3 \\
\textbf{Ours} & \textbf{100.0}$\pm$0.0 & 99.2$\pm$0.3 & \textbf{99.5}$\pm$0.5 & \textbf{97.0}$\pm$1.5 & \textbf{98.9}$\pm$0.4 \\
\textit{$\Delta$ (Ours $-$ Base)} & \textit{+0.7} & \textit{0.0} & \textit{+1.1} & \textit{+4.1} & \textit{+1.5} \\
\bottomrule
\end{tabular}}
\caption{LIBERO success rates. Matched GR00T-N1.6 entries are mean $\pm$ SD over three distinct training-seed checkpoints (100 episodes per suite each). Non-GR00T values follow Spatial Forcing~\cite{li2025spatialforcing}; $\Delta$ is not ranked.}
\pdfanchorlabel{tab:libero_results}
\end{table*}

On standard LIBERO (\Tabref{tab:libero_results}), three checkpoints trained with distinct random seeds are each evaluated; the matched average rises from 97.4\% to 98.9\%, demonstrating that constructing and injecting the multi-view camera field improves downstream action performance. The largest gain ($+4.1$ points) occurs on long-horizon LIBERO-10, where depth ambiguity, cross-view occlusion, and spatial errors can accumulate across multiple subgoals. Performance is maintained or improved on the shorter suites, showing that geometry shaping preserves the base policy's already strong in-domain behavior.

\paragraph{2) LIBERO-Plus robustness.}
LIBERO-Plus~\cite{fei2025liberoplus} evaluates the same manipulation tasks under seven predefined perturbation families spanning camera and layout shifts, appearance changes, noise, and language variation.
\begin{table*}[t!]
\centering
{\small
\setlength{\tabcolsep}{3.9pt}
\renewcommand{\arraystretch}{0.96}
\begin{tabular}{lcccccccc}
\toprule
Method & Camera & Robot & Language & Light & Background & Noise & Layout & Total \\
\midrule
\rowcolor{tablegray}
\multicolumn{9}{c}{\textbf{Published/Concurrent VLA baselines}} \\
$\pi_0$ \cite{black2025pi0} & 13.8 & 6.0 & 58.8 & 85.0 & 81.4 & \textbf{79.0} & 68.9 & 53.6 \\
OpenVLA-OFT \cite{kim2025oft} & \textbf{56.4} & 31.9 & \textbf{79.5} & \underline{88.7} & \textbf{93.3} & 75.8 & \underline{74.2} & \textbf{69.6} \\
UniVLA \cite{bu2025latentactions} & 1.8 & 46.2 & 69.6 & 69.0 & 81.0 & 21.2 & 31.9 & 42.9 \\
Evo-Depth$^{\dagger}$ \cite{lin2026evodepth} & \underline{47.2} & \underline{49.2} & \underline{78.9} & 88.1 & 76.4 & \underline{77.6} & 69.6 & \textbf{69.6} \\
\midrule
\rowcolor{tablegray}
\multicolumn{9}{c}{\textbf{GR00T-based comparison}} \\
GR00T-N1.6 (Base) \cite{nvidia2025gr00tn16} & 20.9 & 40.2 & 35.0 & 65.4 & 76.3 & 27.8 & 51.3 & 42.8 \\
\textbf{Ours} & 24.1 & \textbf{54.1} & 72.9 & \textbf{95.1} & \underline{91.8} & 61.3 & \textbf{74.9} & \underline{65.2} \\
\textit{$\Delta$ (Ours $-$ Base)} & \textit{+3.2} & \textit{+13.9} & \textit{+37.9} & \textit{+29.7} & \textit{+15.5} & \textit{+33.5} & \textit{+23.6} & \textit{+22.4} \\
\bottomrule
\end{tabular}}
\caption{LIBERO-Plus robustness. Ours gains $+22.4$ points over matched GR00T-N1.6 and improves every category. Other values follow LIBERO-Plus~\cite{fei2025liberoplus} and Evo-Depth~\cite{lin2026evodepth}; $\dagger$ denotes concurrent work.}
\pdfanchorlabel{tab:libero_plus_results}
\end{table*}

On LIBERO-Plus (\Tabref{tab:libero_plus_results}), we evaluate the seed-0 terminal checkpoint zero-shot, without retraining or adaptation, under the same RGB-only inference graph and action recipe. Ours improves Base by 22.4 points on average and performs better across all seven categories, demonstrating that the learned multi-view camera field transfers across diverse perturbations rather than benefiting from target-benchmark checkpoint tuning. Notably, geometry injection implicitly incorporates the training-time camera parameters into the learned geometric representation; this dependence limits the improvement under camera-viewpoint and field-of-view perturbations.

\paragraph{3) RoboTwin evaluation across action families.}
Before training, candidate RoboTwin tasks are grouped by their dominant action primitive---brief target activation (Touch), grasp--transport--placement (Move and Place), or sustained force and articulated interaction (Contact)---and two representatives per family form a balanced six-task suite. Each task contributes 50 demonstration episodes (300 in total), and the suite is jointly trained for 180k steps. These tasks require joint localization of the end effector, object, and target under occlusion and changing visibility. Across 1200 rollouts per policy (\Tabref{tab:robotwin_results}), success rises from 38.6\% for Base (95\% Wilson CI: 35.9--41.4\%) to 61.9\% for Ours (59.1--64.6\%); all six tasks improve. Family gains are $+11.0$ for Touch, $+30.5$ for Move and Place, and $+28.5$ for Contact.

\begin{table}[t]
\centering
{\small
\setlength{\tabcolsep}{3.2pt}
\renewcommand{\arraystretch}{0.92}
\begin{tabular}{llrrr}
\toprule
Family & Task & Base & Ours & $\Delta$ \\
\midrule
Touch & Alarm clock & 92.0 & \textbf{99.5} & \textit{+7.5} \\
      & Bell        & 85.5 & \textbf{100.0} & \textit{+14.5} \\
Move/Place & Pill bottle & 6.5 & \textbf{40.5} & \textit{+34.0} \\
      & Phone stand & 0.0 & \textbf{27.0} & \textit{+27.0} \\
Contact & Stapler   & 43.0 & \textbf{93.5} & \textit{+50.5} \\
        & Switch    & 4.5 & \textbf{11.0} & \textit{+6.5} \\
\midrule
\multicolumn{2}{l}{Overall} & 38.6 & \textbf{61.9} & \textit{+23.3} \\
\bottomrule
\end{tabular}}
\caption{RoboTwin results over two rollout-seeded rounds (200 rollouts per task); the six-task evaluation suite was pre-specified before training.}
\pdfanchorlabel{tab:robotwin_results}
\end{table}

\subsection{Ablation study}
On LIBERO-10, three independently trained seed checkpoints show average gains of $+1.12$ points for Depth only and $+1.33$ for Cross-view only, while Full achieves the largest gain of $+4.12$ points (\Tabref{tab:libero10_ablation}).
\begin{table}[t]
\centering
{\small
\setlength{\tabcolsep}{1.5pt}
\renewcommand{\arraystretch}{0.96}
\begin{tabular}{lcccccc}
\toprule
Configuration & Depth & CV & R1 & R2 & R3 & Mean $\pm$ SD \\
\midrule
Native Base & -- & -- & 92.10 & 92.52 & 94.02 & 92.88 $\pm$ 1.01 \\
Depth only & \checkmark & -- & 93.00 & 94.00 & 95.00 & 94.00 $\pm$ 1.00 \\
Cross-view only & -- & \checkmark & 95.05 & 91.50 & 96.07 & 94.21 $\pm$ 2.40 \\
\textbf{Full} & \checkmark & \checkmark & 97.00 & 98.50 & 95.50 & \textbf{97.00 $\pm$ 1.50} \\
\bottomrule
\end{tabular}}
\caption{LIBERO-10 ablation with identical action data and 60k-step training; R1--R3 are distinct training-seed checkpoints evaluated once each.}
\pdfanchorlabel{tab:libero10_ablation}
\end{table}
These results support the complementarity of the two objectives: each improves the base policy on average, while their joint use produces the largest mean gain.

\subsection{Real-robot experiments}
\begin{figure}[t]
\centering
\includegraphics[width=\linewidth]{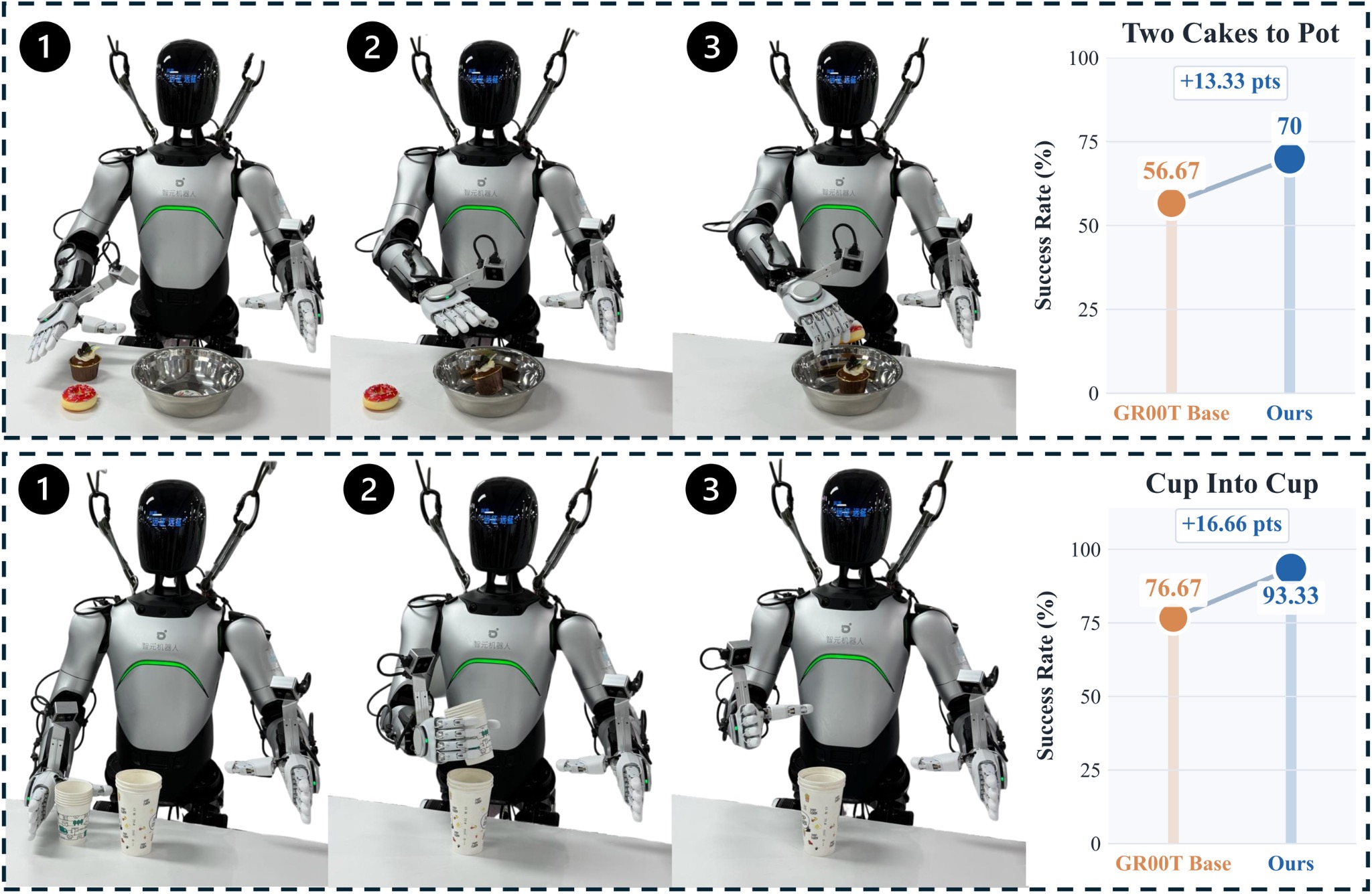}
\caption{Real-robot RGB-only pilot on Agibot Expedition A2 (30 trials/policy/task); inference uses no geometry or auxiliary head.}
\pdfanchorlabel{fig:real_world_results}
\end{figure}
Using 100 teleoperated demonstrations per task, we train Base and Ours on two tasks---two-cake placement and cup nesting---and evaluate 30 trials per task (\Figref{fig:real_world_results}). Ours succeeds in 49/60 trials (81.7\%; 95\% Wilson CI: 70.1--89.4\%) versus 40/60 for Base (66.7\%; 54.1--77.3\%), improving both tasks (17/30$\to$21/30 and 23/30$\to$28/30). Despite overlapping intervals, the consistent direction provides pilot-scale physical feasibility evidence for geometry-shaped RGB-only deployment.

\section{Conclusion}
\label{sec:conclusion}
Multi-View Unified Camera Fields (MVUCF) injects metric-depth recoverability and same-world-point consistency into the action-facing backbone before policy learning. Held-out probes verify both properties; matched simulations improve across LIBERO, LIBERO-Plus, and RoboTwin, and the physical pilot shows the same favorable direction. Removing geometry inputs, calibration, and auxiliary heads before action training preserves native RGB-only inference.

\paragraph{Scope and outlook.}
MVUCF assumes accurate training-time calibration. Robustness to corrupted or noisy calibration labels remains untested and motivates calibration-noise-robust, uncertainty-aware target construction.

\input{appendix}

\bibliography{references}

\end{document}

%% file: appendix.tex
\clearpage
\appendix
\setcounter{secnumdepth}{1}

\section{Training Setup and Notation}
\label{app:training}

A synchronized multi-camera sample contains RGB images $\{I_i\}_{i=1}^{V}$, depth maps $\{D_i\}_{i=1}^{V}$, intrinsics $\{\mathbf K_i\}_{i=1}^{V}$, and camera-to-world poses $\{\mathbf T_{i\to w}\}_{i=1}^{V}$. We denote the inverse world-to-camera transform by $\mathbf T_{w\to i}=\mathbf T_{i\to w}^{-1}$. The VLA backbone produces a hidden grid
\begin{equation}
    \mathbf F_i \in \mathbb R^{H_g \times W_g \times C}
\end{equation}
for each view $i$ at the action-facing layer. A coordinate-query depth head $g_\phi$ predicts metric depth at sampled continuous grid coordinates from $\mathbf F_i$, while a projector $h_\omega$ maps hidden tokens to normalized embeddings for correspondence matching.

The training recipe has two stages. First, geometry injection updates only VLM layers 8--15 and the two auxiliary heads using depth and cross-view losses; the visual encoder and layers 0--7 remain frozen. The update range is fixed across datasets and is not selected using downstream benchmark results. Second, the auxiliary heads are removed and the action module is trained on the geometry-injected backbone. The second stage is intentionally identical to the native action-training recipe except that the hidden states have been reshaped by the first stage. We use a global batch size of 128 throughout. Geometry injection runs for 50k optimization steps. Action training runs for 60k steps on LIBERO. For RoboTwin, the six-task suite is jointly trained for 180k steps under balanced task sampling, corresponding to 30k task-equivalent steps per task. We always use the final checkpoint at the predeclared horizon (\texttt{checkpoint-50000}, \texttt{checkpoint-60000}, and \texttt{checkpoint-180000}, respectively), without validation- or downstream-success-based checkpoint selection. LIBERO experiments use two NVIDIA H100 GPUs, whereas RoboTwin experiments use four NVIDIA H100 GPUs. Both native and geometry-injected variants use AdamW (\texttt{adamw\_torch}) with initial learning rate $10^{-4}$, weight decay $10^{-5}$, and linear warmup over the first 5\% of steps, matching the official GR00T-N1.6 finetuning defaults. Under these hardware configurations, each geometry-injection run takes approximately 18 hours. Action-training duration varies with the benchmark-specific schedule, so we do not report a single wall-clock value. Geometry injection therefore adds offline training cost but adds no parameters or FLOPs to the final deployment graph.

\section{Preprocessing-Aware Coordinate Mapping}
\label{app:coordinates}

The auxiliary labels are generated from raw calibrated camera geometry, while the losses are applied on the VLM token grid. We therefore distinguish the raw camera image plane from the final token grid and define a composed coordinate transform rather than using a single resize ratio.

For a source raw pixel $\mathbf u_i^{\mathrm{raw}}$ with valid depth $z_i^\ast$, intrinsics $\mathbf K_i$, and camera-to-world pose $\mathbf T_{i\to w}$, reprojection is performed entirely in the raw camera coordinate system:
\begin{equation}
\label{eq:supp_raw_reproject}
\begin{split}
\mathbf P_w &= \mathbf T_{i\to w}\,\pi^{-1}(\mathbf u_i^{\mathrm{raw}},z_i^\ast,\mathbf K_i),\\
\mathbf u_j^{\mathrm{raw}} &= \pi(\mathbf T_{w\to j}\mathbf P_w,\mathbf K_j).
\end{split}
\end{equation}
Only after obtaining the target raw pixel do we map it into the VLM token grid:
\begin{equation}
\label{eq:supp_composed_transform}
    \mathbf q_j^\ast = T_{\mathrm{img}\to\mathrm{grid}}(\mathbf u_j^{\mathrm{raw}}).
\end{equation}
The composed transform used in our implementation is
\begin{equation}
\begin{aligned}
T_{\mathrm{img}\to\mathrm{grid}}
&=T_{\mathrm{unshuffle}}^{(2)}\circ T_{\mathrm{patch}}^{(14)}\circ T_{\mathrm{smart}} \\
&\quad \circ T_{\mathrm{resize2}}\circ T_{\mathrm{crop}}\circ T_{\mathrm{resize1}} \\
&\quad \circ T_{\mathrm{letterbox}}.
\end{aligned}
\end{equation}
It corresponds to the following chain: raw image pixel $\rightarrow$ letterbox-to-square $\rightarrow$ SmallestMaxSize on the shortest edge $\rightarrow$ fractional crop $\rightarrow$ SmallestMaxSize again $\rightarrow$ Eagle smart-resize $\rightarrow$ SigLIP patch grid with patch size 14 $\rightarrow$ pixel-unshuffle factor 2 $\rightarrow$ final token grid. This is the transform implemented by \texttt{compose\_orig\_uv\_to\_grid\_xy}. Thus, cross-view labels are not generated by naively scaling raw pixels to the token grid.

\begin{algorithm}[t]
\caption{Geometry-injection training}
\label{alg:geometry_injection}
{\small
\begin{algorithmic}[1]
\REQUIRE Dataset $\mathcal D$ with RGB, depth, intrinsics, extrinsics; VLA backbone $f_\theta$; depth head $g_\phi$; correspondence projector $h_\omega$.
\FOR{each minibatch $B \sim \mathcal D$}
    \STATE Encode all RGB views and read the action-facing hidden grids $\{\mathbf F_i\}$ from $f_\theta$.
    \STATE For every sample/view, draw 3072 valid base queries: 1536 uniform and 1536 importance-sampled toward near depth and large depth gradients.
    \STATE Predict metric depth $\hat z_i(\mathbf q)=g_\phi(\mathbf F_i,\mathbf q)$ for every $\mathbf q\in\mathcal Q_i$.
    \STATE Construct query-level targets and masks through the composed image-to-grid transform; evaluate losses only at valid depth locations.
    \STATE Construct cross-view positives by geometric back-projection, target-view re-projection, and visibility filtering.
    \STATE Compute
    \[
        \mathcal L = \lambda_{\mathrm{depth}}\mathcal L_{\mathrm{depth}} + \lambda_{\mathrm{cv}}\mathcal L_{\mathrm{cv}} .
    \]
    \STATE Update the trainable VLM layers and the auxiliary heads.
\ENDFOR
\STATE Discard $g_\phi$ and $h_\omega$.
\STATE Train the downstream action module using the same RGB-only inputs and inference graph as the native policy.
\end{algorithmic}}
\end{algorithm}

\section{Coordinate-Based Metric Grounding}
\label{app:metric_grounding}

The final runs use \texttt{depth\_decoder\_type=coord} and \texttt{coord\_supervision\_mode=query}; they do not supervise a dense full-image prediction at every training step. The head taps the post-block layer-15 grid with hidden width 2048. For a continuous token-grid location $\mathbf q=(x,y)$,
\begin{equation}
    (\hat z_i(\mathbf q),\hat s_i(\mathbf q))=g_\phi(\mathbf F_i,\mathbf q),
\end{equation}
where $\hat z>0$ is metric depth and $\hat s$ is predicted log variance.

Before continuous sampling, the hidden grid is passed through a GridTrunk:
\begin{equation}
\begin{aligned}
&\mathrm{LayerNorm}(2048) \to \mathrm{Linear}(2048,1024) \\
&\to \mathrm{GELU} \to \mathrm{Reshape} \\
&\to \mathrm{Conv}_{3\times 3}(1024,1024;s{=}1,p{=}1) \to \mathrm{GELU}.
\end{aligned}
\end{equation}
For sample $b$ and view $i$, sampling coordinates are normalized by
\begin{equation}
    \tilde{x}_{bi}=2x/(W_{g,bi}-1)-1,\qquad
    \tilde{y}_{bi}=2y/(H_{g,bi}-1)-1.
\end{equation}
The convolution preserves the valid grid and performs no downsampling. Features are queried with \texttt{grid\_sample} using \texttt{mode=bilinear}, \texttt{align\_corners=True}, and \texttt{padding\_mode=border}. The sub-cell phase is
\begin{equation}
\boldsymbol\varphi(\mathbf q)=(x-\lfloor x\rfloor,\;y-\lfloor y\rfloor).
\end{equation}
With \texttt{coord\_corner\_features=false} and \texttt{local\_context=true}, the query feature is
\begin{equation}
\mathbf r(\mathbf q)=
[\mathbf f(\mathbf q);\Delta_x\mathbf f(\mathbf q);\Delta_y\mathbf f(\mathbf q);\boldsymbol\varphi(\mathbf q)]\in\mathbb R^{3074},
\end{equation}
where each sampled or local-difference feature is 1024-D. Absolute image position is not provided. The two prediction branches are
\begin{equation}
\begin{aligned}
\hat z &: 3074\to1024\to1024\to1, &&\mathrm{GELU},\mathrm{GELU},\mathrm{Softplus},\\
\hat s &: 3074\to1024\to1, &&\mathrm{GELU},\mathrm{clamp}[-6,6].
\end{aligned}
\end{equation}
The same module can emit dense evaluation maps $(\texttt{depth\_pred},\texttt{log\_var\_dense})$, but training uses the query outputs $(\texttt{depth\_query},\texttt{log\_var\_query})$.

For the default $9\times9$ token grid, the coordinate head defines a $144\times144$ continuous query lattice using dense-query factor $P=16$. Each sample/view draws 3072 base queries: 50\% uniformly from valid locations and 50\% by importance sampling favoring near surfaces (inverse depth) and depth edges. The sampler also requests the $+x$ and $+y$ neighbors used by the metric gradient term. Targets are read by nearest-neighbor mapping after the inverse image-to-grid transform, and every term is masked by valid depth in the range $[0.05,5.0]$ m.

The query-mode objective is
\begin{equation}
\label{eq:depth_loss}
\begin{split}
\mathcal L_{\mathrm{depth}}={}&0.5\mathcal L_{\mathrm{silog}}+1.25\mathcal L_{\mathrm{inv}}+1.0\mathcal L_{\mathrm{grad}}\\
&+0.1\mathcal L_{\mathrm{seam}}+0.05\mathcal L_{\mathrm{unc}}.
\end{split}
\end{equation}
The outer depth-ground-truth weight is 1.0 and there is no depth-head warmup. The SILog branch uses adaptive per-view weighting (\texttt{depth\_silog\_adaptive\_view\_weight=true}). The inverse-depth Smooth-$L_1$ weight is fixed at 1.25 in query mode; the dense-branch final/ramp and foreground-weighting settings are inactive for these runs. The uncertainty term consumes the clamped query log variance. Both auxiliary terms are active in the query-mode implementation; we report their code-level names and weights to distinguish them from the inactive dense-only options.

The implemented gradient term is metric depth first-difference matching:
\begin{equation}
\label{eq:supp_grad}
\begin{split}
\mathcal{L}_{\mathrm{grad}}
=\frac{1}{|\mathcal{D}|}
\sum_{\Delta}\sum_{\mathbf q\in\mathcal{D}}
\bigl|(\hat z_{\mathbf q+\Delta}-\hat z_{\mathbf q})\\
-(z^\ast_{\mathbf q+\Delta}-z^\ast_{\mathbf q})\bigr|,
\end{split}
\end{equation}
where $\Delta\in\{(1,0),(0,1)\}$ and $\mathcal D$ contains valid neighboring depth pairs. This term preserves local metric structure and discourages constant-depth collapse.

\section{Cross-View Target Construction and Matching}
\label{app:crossview}
\subsection{Geometric target construction}

For each source view $s$ and target view $t$, positives are defined by raw-frame geometric reprojection rather than by visual similarity. Eq.~\eqref{eq:supp_raw_reproject} maps a source raw pixel and its ground-truth depth to a world point and then to a target raw pixel. A projected pair is kept only if it satisfies four checks: the target pixel lies inside the raw target image, the target-frame depth is positive, the target depth map is valid at the projected location, and the z-buffer visibility check
\begin{equation}
    |Z_t-z_t^*(\mathbf u_t^{\mathrm{raw}})|<\delta_{\mathrm{vis}}
\end{equation}
is satisfied. We use $\delta_{\mathrm{vis}}=0.10$ m. This value is an empirically selected operating point between target precision and target density. Tighter thresholds discard many geometrically valid pairs because of depth quantization, completion residuals, and sensor noise; looser thresholds increasingly admit ambiguous projections near occlusion boundaries. Our target-quality validation found that 10\,cm substantially expands the pool of valid positive pairs while keeping erroneous matches limited. Accepted target raw pixels are mapped to token-grid coordinates through $T_{\mathrm{img}\to\mathrm{grid}}$ in Eq.~\eqref{eq:supp_composed_transform}. Target construction uses synchronized views with calibrated intrinsics and extrinsics. Calibration generates geometry-injection targets only and is absent from action training and deployment. The reported experiments therefore characterize the standard calibrated target-construction setting used by our data pipeline. Moving a deployment camera instead constitutes viewpoint-distribution shift, because the deployed policy consumes no camera matrix.

\begin{algorithm}[t]
\caption{Cross-view label construction}
\label{alg:cv_labels}
{\small
\begin{algorithmic}[1]
\REQUIRE Source view $s$, target view $t$, raw depth maps, intrinsics, camera poses, and $T_{\mathrm{img}\to\mathrm{grid}}$.
\FOR{each sampled valid source raw pixel $\mathbf u_s^{\mathrm{raw}}$}
    \STATE Back-project $\mathbf u_s^{\mathrm{raw}}$ to $\mathbf P_w$ with ground-truth depth and camera intrinsics.
    \STATE Re-project $\mathbf P_w$ into the target raw camera frame to obtain $\mathbf u_t^{\mathrm{raw}}$.
    \STATE Reject the pair if image-bound, depth-validity, positive-depth, or z-buffer checks fail.
    \STATE Map $\mathbf u_t^{\mathrm{raw}}$ to the target token-grid center $\mathbf q_t^\ast=T_{\mathrm{img}\to\mathrm{grid}}(\mathbf u_t^{\mathrm{raw}})$.
    \STATE Use $\mathbf q_t^\ast$ as the center of a Gaussian soft label on the target token grid.
    \STATE Sample hard negatives from spatial rings, in-batch tokens, and alternate views.
\ENDFOR
\RETURN Positive centers, soft labels, and hard-negative sets.
\end{algorithmic}}
\end{algorithm}

\subsection{Matching objective}

The shared cross-view projector is deliberately minimal:
\begin{equation}
 h_\omega(\mathbf f)=\operatorname{normalize}_2\!\left(\mathbf W\,\operatorname{LN}_{2048}(\mathbf f)\right),
 \qquad \mathbf W\in\mathbb R^{128\times2048},
\end{equation}
with no hidden layer or activation. L2 normalization uses $\epsilon=10^{-6}$. Thus the projector cannot create a deep auxiliary representation; the VLM hidden state itself must become linearly correspondence-readable. For global matching, the pairwise logit is
\begin{equation}
    \ell_{mn}=\tau (\mathbf e_m^s)^\top \mathbf e_n^t,\qquad \tau=10.0.
\end{equation}
Here $\tau$ is a fixed non-learnable multiplicative logit scale rather than a divisor. Instead of a one-hot target, we use a Gaussian soft label centered at the geometrically projected token $j^*(m)$:
\begin{equation}
    y_{mn} \propto \exp\left(-\frac{\|\mathbf c_n-\mathbf c_{j^*(m)}\|_2^2}{2\sigma^2}\right),
\end{equation}
where $\mathbf c_n$ is the target-grid coordinate and $\sigma=0.75$. The full correspondence loss is
\begin{equation}
\label{eq:cv_loss}
    \mathcal L_{\mathrm{cv}}=\mathcal L_{\mathrm{softCE}}+
    \lambda_{\mathrm{nce}}\mathcal L_{\mathrm{hnce}}+
    \lambda_{\mathrm{margin}}\mathcal L_{\mathrm{margin}}.
\end{equation}
The soft-target cross-entropy teaches the full target-view similarity distribution around the geometric projection. Let $p_m=j^*(m)$ be the projected positive index and $\mathcal N_m$ the sampled hard-negative set (spatial-ring, in-batch, and alternate-view negatives). The two auxiliary terms are
\begin{equation}
\label{eq:supp_hnce}
\mathcal L_{\mathrm{hnce}}=-\frac1M\sum_{m=1}^{M}\log\frac{\exp(\ell_{mp_m})}{\exp(\ell_{mp_m})+\sum_{n\in\mathcal N_m}\exp(\ell_{mn})},
\end{equation}
\begin{equation}
\label{eq:supp_margin}
\mathcal L_{\mathrm{margin}}=\frac1M\sum_{m=1}^{M}\left[\gamma-\ell_{mp_m}+\max_{n\in\mathcal N_m}\ell_{mn}\right]_+.
\end{equation}
Thus InfoNCE contrasts the projected positive with selected difficult negatives, while the hinge term explicitly enforces a $\gamma$ logit gap from the hardest non-positive. The main runs enable both global matching and local refinement. The local branch searches a $3\times3$ target window; its multiplicative logit scale starts at 50, reaches 30 after an 8k-step anneal, and has no warmup. We use $\lambda_{\mathrm{nce}}=0.1$, $\gamma=0.2$, and $\lambda_{\mathrm{margin}}=0.1$. The outer cross-view loss weight decays linearly from 1.0 to 0.1 over 40k geometry-injection steps, while the outer depth weight remains 1.0. All geometry weights, query counts, and schedules are fixed across datasets rather than tuned per benchmark.

\section{Diagnostic Extraction Protocols}
\label{app:diagnostics}
\subsection{Layer-wise external-geometry diagnostic}

For each geometrically accepted source--target pair, the source feature retrieves all target-view tokens by cosine similarity. We report the median percentile rank of the projected target token (lower is better) and Cohen's $d$ between true-match and randomly sampled non-correspondence similarities (higher is better). The same calibrated correspondences and metric definitions are used at every probed layer; only the extracted VGGT layer changes. Shallow features preserve local patch identity and separate true matches from random pairs, whereas deeper features become less discriminative after global attention mixing. A deep geometry representation can therefore retain useful 3D abstraction without providing the localized same-point correspondence space consumed by the action module.

\subsection{Action-to-attention visualization}

The action-to-attention figure in the main text is a qualitative diagnostic rather than a training component or quantitative endpoint. It asks whether the representation change induced by geometry injection is visually reflected in the action-facing computation during a rollout. We evaluate the native policy and the geometry-injected policy on the same task family and visualize spatial relevance at matched execution steps; no bounding-box attention ratio or calibrated cross-policy magnitude is inferred from these displays.

\begin{algorithm}[t]
\caption{Action-to-attention diagnostic}
\label{alg:attention}
{\small
\begin{algorithmic}[1]
\REQUIRE Trained policy, evaluation rollout, selected steps $\{t_k\}$.
\FOR{each selected step $t_k$}
    \STATE Run a forward pass using the RGB observations available at $t_k$.
    \STATE Collect the action-facing attention maps using the configuration specified below.
    \STATE Aggregate the selected attention maps to obtain a token-level relevance map.
    \STATE Resize the token relevance map to the image plane and normalize it for display.
    \STATE Overlay the heatmap on the corresponding RGB observation.
\ENDFOR
\STATE Compare native and geometry-injected policies at the same task steps.
\end{algorithmic}}
\end{algorithm}

The maps are extracted from image cross-attention blocks in the DiT action head, in the direction action-query tokens $\rightarrow$ image key/value tokens. With the default \texttt{attend\_text\_every\_n\_blocks=2}, we retain even cross-attention blocks that are not text cross-attention blocks, equivalently indices satisfying $\texttt{idx}\bmod 4=2$; for a 16-layer DiT these are blocks $\{2,6,10,14\}$. We use all attention heads and take a simple mean over selected layers, heads, action queries, and denoising steps. There is no attention rollout or gradient weighting. Each displayed cell is independently min--max normalized before overlay, so color intensity should be compared spatially within a cell rather than numerically across cells.

This diagnostic is qualitative and does not replace success-rate evaluation or representation probing.

\section{Evaluation Protocol Details}
\label{app:evaluation}
\subsection{LIBERO multi-seed evaluation and ablation}

The standard LIBERO comparison uses three independently trained checkpoints from distinct training seeds for each method. Each checkpoint is evaluated once with 100 episodes per suite, implemented as 20 episodes in each of five parallel environments. Thus, the reported variation is across training seeds rather than repeated evaluations of one checkpoint. LIBERO-Plus uses the seed-0 terminal checkpoint for zero-shot evaluation, without retraining or adaptation.

\begin{table}[t]
\centering
\small
\setlength{\tabcolsep}{3.0pt}
\begin{tabular}{@{}llccc@{}}
\toprule
Method & Suite & Ckpt 1 & Ckpt 2 & Ckpt 3 \\
\midrule
Base & Spatial & 98.00 & 100.00 & 100.00 \\
Ours & Spatial & 100.00 & 100.00 & 100.00 \\
Base & Object & 99.00 & 100.00 & 98.50 \\
Ours & Object & 99.00 & 99.00 & 99.50 \\
Base & Goal & 99.50 & 98.55 & 97.07 \\
Ours & Goal & 100.00 & 99.50 & 99.00 \\
Base & LIBERO-10 & 92.10 & 92.52 & 94.02 \\
Ours & LIBERO-10 & 97.00 & 98.50 & 95.50 \\
\bottomrule
\end{tabular}
\caption{Per-checkpoint success rates (\%) used for the standard LIBERO averages. Each column is one independently trained checkpoint from a distinct training seed and one 100-episode evaluation per suite.}
\label{tab:supp_libero_rounds}
\end{table}

For the component ablation, the depth-only configuration is a direct metric-depth auxiliary baseline without cross-view reprojection. It obtains 93.00\%, 94.00\%, and 95.00\% across three distinct training-seed checkpoints, while the cross-view-only configuration obtains 95.05\%, 91.50\%, and 96.07\%. Their means are 94.00\% and 94.21\%, respectively, compared with 92.88\% for the native checkpoint and 97.00\% for the full geometry-injected checkpoint. All four configurations use the same 60k-step action-training budget and one evaluation per checkpoint.

\subsection{RoboTwin evaluation scope}

The main paper reports a taxonomy-driven six-task RoboTwin diagnostic suite rather than a full-benchmark leaderboard. Before either policy is evaluated, candidate tasks are assigned by one dominant action primitive: brief target activation (Touch), grasp--transport--placement (Move and Place), or sustained force/articulated interaction (Contact). We then fix two tasks whose dominant primitive is unambiguous in each family. The task list is therefore determined from action semantics and balanced family coverage, not from the observed gains of either policy. \emph{Touch} denotes brief target activation with little object transport and includes \texttt{click\_alarmclock} and \texttt{click\_bell}. \emph{Move and Place} denotes grasp--transport--placement behavior and includes \texttt{move\_pillbottle\_pad} and \texttt{place\_phone\_stand}. \emph{Contact} denotes sustained force or articulated interaction and includes \texttt{press\_stapler} and \texttt{turn\_switch}. This balanced construction is intended to diagnose performance across action families, not to claim full-benchmark coverage. The six tasks are jointly trained in a single policy for 180k steps under balanced task sampling (30k task-equivalent steps per task). Each table entry averages two evaluation rounds with independently seeded rollouts and 100 rollouts per task per round. The rounds reuse the same trained checkpoint and therefore should not be interpreted as independent training seeds. Because two rounds do not support a stable estimate of between-round variance, Table~\ref{tab:supp_robotwin_ci} reports aggregate binomial uncertainty over 200 rollouts per task.

\begin{table*}[t]
\centering
\small
\setlength{\tabcolsep}{4.0pt}
\begin{tabular}{lrrrrrrr}
\toprule
Method & Alarm & Bell & Pill bottle & Phone stand & Stapler & Switch & Overall \\
\midrule
Base successes & 184/200 & 171/200 & 13/200 & 0/200 & 86/200 & 9/200 & 463/1200 \\
Base 95\% CI & 87.4--95.0 & 80.0--89.7 & 3.8--10.8 & 0.0--1.9 & 36.3--49.9 & 2.4--8.3 & 35.9--41.4 \\
Ours successes & 199/200 & 200/200 & 81/200 & 54/200 & 187/200 & 22/200 & 743/1200 \\
Ours 95\% CI & 97.2--99.9 & 98.1--100.0 & 33.9--47.4 & 21.3--33.5 & 89.2--96.2 & 7.4--16.1 & 59.1--64.6 \\
\bottomrule
\end{tabular}
\caption{RoboTwin success counts and 95\% Wilson intervals. They quantify rollout-level binomial uncertainty for a fixed checkpoint, not retraining variability.}
\label{tab:supp_robotwin_ci}
\end{table*}

\section{Architectures and Hyperparameters}
\label{app:hyperparameters}

Table~\ref{tab:training_recipe} reports the optimization budget and hardware used for each benchmark. Table~\ref{tab:hparams} summarizes the geometry-specific settings. Optimizer settings are identical for native and geometry-injected comparisons; the tables enumerate the GR00T-N1.6 defaults and every setting changed by our method.

\begin{table}[t]
\centering
\small
\setlength{\tabcolsep}{2.1pt}
\begin{tabular}{@{}p{0.47\columnwidth}ccc@{}}
\toprule
\rowcolor{tablegray}
\textbf{Stage / benchmark} & \textbf{Steps} & \textbf{BS} & \textbf{Hardware} \\
\midrule
Geo / LIBERO ($\sim$18 h) & 50k & 128 & 2$\times$H100 \\
Action / LIBERO & 60k & 128 & 2$\times$H100 \\
Geo / RoboTwin ($\sim$18 h) & 50k & 128 & 4$\times$H100 \\
Action / RoboTwin (joint) & 180k & 128 & 4$\times$H100 \\
\bottomrule
\end{tabular}
\caption{Training schedules, global batch sizes, and hardware. Approximate wall-clock time is reported only for geometry injection because action-training duration varies across benchmark-specific schedules. We use terminal checkpoints at 50k geometry steps, 60k LIBERO action steps, and 180k joint RoboTwin action steps, without validation- or downstream-success-based selection.}
\label{tab:training_recipe}
\end{table}

\begin{table*}[t]
\centering
\small
\begin{tabular}{@{}p{0.22\textwidth}p{0.73\textwidth}@{}}
\toprule
\rowcolor{tablegray}
\textbf{Module} & \textbf{Exact architecture / active mode} \\
\midrule
Tap and update range & Post-block Layer 15, hidden dim 2048; update Layers 8--15 \\
Depth GridTrunk & LN(2048)$\to$Linear(1024)$\to$GELU$\to$Conv$_{3\times3}$(1024)$\to$GELU \\
Depth query input & sampled (1024) + local-$x$ (1024) + local-$y$ (1024) + phase (2) = 3074 \\
Depth branch & Linear(3074,1024)$\to$GELU$\to$Linear(1024,1024)$\to$GELU$\to$Linear(1024,1)$\to$Softplus \\
Log-variance branch & Linear(3074,1024)$\to$GELU$\to$Linear(1024,1)$\to$clamp $[-6,6]$ \\
CV projector & LN(2048) $\rightarrow$ Linear(2048,128) $\rightarrow$ $\ell_2$-normalization ($\epsilon=10^{-6}$); no hidden layer or activation \\
\bottomrule
\end{tabular}
\caption{Auxiliary-head architectures. The final depth configuration is coordinate-query supervision, not dense full-image supervision; both heads are discarded before action training.}
\label{tab:aux_arch}
\end{table*}

\begin{table*}[t]
\centering
\small
\begin{tabular}{@{}p{0.28\textwidth}p{0.67\textwidth}@{}}
\toprule
\rowcolor{tablegray}
\textbf{Item} & \textbf{Value} \\
\midrule
Optimizer; LR / WD; warmup & AdamW; $10^{-4}$ / $10^{-5}$; 0.05 \\
Depth decoder / supervision & \texttt{coord} / \texttt{query}; no warmup \\
Depth layer / MLP hidden & 15 / 1024 \\
Depth patch size / image edge & 28 / 512; original chain 256$\times$256 \\
Dense query factor / lattice & $P=16$ / $9\times9\to144\times144$ \\
Query budget / mixture & 3072 per view; 50\% uniform, 50\% importance \\
Depth valid range & 0.05--5.0 m \\
$\lambda_{\rm depth}$ & 1.0 \\
SILog / inverse / gradient & 0.5 / 1.25 / 1.0 \\
Seam / uncertainty & 0.1 / 0.05 \\
Adaptive SILog view weights & Enabled \\
CV global / local refinement & Enabled / enabled \\
CV output dimension & 128 \\
Global CV logit scale & 10.0, fixed \\
Local window / scale & $3\times3$; 50$\to$30 over 8k, no warmup \\
Outer CV loss weight & 1.0$\to$0.1 over 40k steps \\
Soft-label $\sigma$ / margin $\gamma$ & 0.75 / 0.2 \\
Hard-NCE / margin weights & 0.1 / 0.1 \\
Visibility threshold & 0.10 m \\
Geometry steps / global batch & 50k / 128 \\
Downstream VLM / deployment & Frozen / no auxiliary modules or FLOPs \\
\bottomrule
\end{tabular}
\caption{Active geometry-injection configuration. Dense-branch inverse-weight ramps and foreground weighting are inactive because the final runs use query supervision. Settings are fixed across datasets.}
\label{tab:hparams}
\end{table*}

\paragraph{Layer choice and front-end portability.}
The trainable range is fixed before downstream evaluation. VLM probing indicated that metric-depth information is concentrated in deeper blocks, motivating supervision near the action interface rather than in early appearance layers. During method development, placing the cross-view objective at layer 12 degraded token-level separation after training; layer 15 was therefore selected as the supervision tap because it retained stronger same-point discrimination and is the final hidden spatial grid directly consumed by the action module. We update the contiguous upper half (layers 8--15) so this signal can propagate through the action-facing blocks while the visual encoder and lower VLM blocks remain frozen. This localizes full-rank representation shaping around the action interface; unlike PEFT, it does not add low-rank or adapter modules. After injection, the VLM is frozen during downstream action training for both Base and Ours. This yields a fixed, probe- and interface-motivated configuration that is reused across all datasets. Likewise, $T_{\mathrm{img}\to\mathrm{grid}}$ is not intrinsically tied to Eagle/SigLIP: it denotes the deterministic coordinate map induced by a front-end. For a standard ViT it reduces to resize/crop and patch-index transforms; architectures without pixel-unshuffle use the identity for that factor, and strided convolutional front-ends can use their cumulative stride and token-center offsets. The raw-camera reprojection, visibility filtering, soft targets, and losses are unchanged.

\paragraph{Visibility-threshold selection and target-noise safeguards.}
The 10\,cm visibility gate is fixed before policy evaluation using target construction itself: we balance retained positive-pair yield against incorrect or ambiguous reprojections, without using downstream success to select the value. Training targets are accepted only on valid depth, positive target-frame depth, in-image projection, and z-buffer consistency. The Gaussian correspondence target tolerates sub-token coordinate shifts, whereas the visibility gate rejects larger depth inconsistencies. For physical data, valid sensor measurements are preserved and completion is applied only to invalid holes; the validity mask remains active in the depth loss. Together, these mechanisms address quantization, missing returns, occlusion boundaries, and small residual alignment errors encountered during target construction.

\section{Assumptions, Data Reuse, and Real-World Depth Processing}
\label{app:scope}

\paragraph{Stage-wise data reuse.}
For both LIBERO and RoboTwin, geometry injection and action training use the same official open-source demonstration trajectories and the same training split. For RoboTwin specifically, each task uses the official \texttt{clean50} split containing 50 demonstration trajectories, yielding 300 trajectories across the six-task suite. We replay these same trajectories to record synchronized depth together with camera intrinsics and extrinsics for geometry-target construction. The replay-derived depth and calibration are used only during geometry injection, whereas downstream action training uses the original RGB-action records from the same trajectories. No additional demonstrations, held-out evaluation trajectories, or task-specific data are introduced for Ours. Thus, Base and Ours differ in the training objective applied before action learning, not in demonstration coverage.

The geometry-injection stage uses temporally synchronized views, calibrated camera parameters, and usable depth on supervised pixels. Calibration is consumed only by the offline target builder. Neither camera matrices nor depth are inputs to downstream action training or deployment, so the final policy retains a standard RGB-only interface. The current study evaluates the calibrated target-construction setting used by our data pipeline; systematic pose corruption is a complementary stress-test direction.

Transparent, reflective, and textureless regions can produce missing or biased depth. Invalid measurements are excluded by the validity mask, and physical-data completion fills only invalid holes while preserving measured depths, improving usable target coverage without altering valid sensor returns. The reported repeated evaluations reuse one trained checkpoint and quantify rollout uncertainty, while LIBERO uses distinct training seeds. Stable Object/Goal control provides task-domain evidence that geometry shaping preserves the semantic capabilities needed by the evaluated manipulation tasks; broader open-domain probing is a complementary extension.

\subsection{Real-world depth completion}

For real-world data, depth maps can contain holes caused by missing sensor returns, reflective surfaces, or occlusion. Valid sensor depths are kept unchanged; only invalid holes are completed and refined before target construction using the publicly downloadable LingBot-Depth code and pretrained weights released with arXiv:2601.17895~\cite{tan2026lingbotdepth}, checkpoint \texttt{robbyant/lingbot-depth-}\allowbreak\texttt{pretrain-vitl-14-v0.5}.

Bilinear sampling in the coordinate-query head applies only to hidden features at continuous token coordinates; it is not used to interpolate supervision. After completion, the depth alignment pipeline strictly uses nearest-neighbor sampling. During data conversion, depth frames are downsampled with \texttt{ffmpeg} using \texttt{flags=neighbor}. During training, the model's depth preprocessing function (\texttt{\_resize\_depth\_gt\_for\_dense}) maps coordinates back to the original image and rounds to the nearest pixel via \texttt{torch.round}, while cropping is applied as a coordinate-space operation without any interpolation smoothing. We separately maintain a valid mask, and the depth loss is computed exclusively on valid pixels to prevent any remaining invalid regions from diffusing into the neighborhood. This nearest-neighbor approach, in contrast to the RGB image path (\texttt{cv2.INTER\_AREA}), ensures that metric depth values at object boundaries are not artificially smoothed or blended between foreground and background. This preprocessing affects only training-target construction. The deployed policy does not receive completed depth, raw depth, camera intrinsics, or camera extrinsics.